# Erkang-Diagnosis-1.1 Technical Report

Jianbing Ma[*], Ao Feng, Zhenjie Gao,Xinyu Song,Li Su, Bin Chen,Wei Wang, Jiamin Wu

Chengdu Lingshu Health Technology Corp. Ltd.
1938193637@qq.com

## Abstract

This report provides a detailed introduction to Erkang-Diagnosis-1.1 model, our company's AI healthcare consulting assistant developed using Alibaba's Qwen-3 model. The Erkang model integrates approximately 500GB of high-quality structured medical knowledge, employing a hybrid approach combining enhanced pre-training and retrieval-enhanced generation to create a secure, reliable, and professional AI health advisor. Through 3-5 efficient interaction rounds, Erkang Diagnosis can accurately understand user symptoms, conduct preliminary analysis, and provide valuable diagnostic suggestions and health guidance. Designed to become users' intelligent health companions, it empowers primary healthcare and health management. To validate, Erkang-Diagnosis-1.1 leads GPT-4 in terms of comprehensive medical exams.

## 1. Introduction

Large Language Models (LLMs) represent a major breakthrough in the field of artificial intelligence, particularly in natural language processing (Vaswani et al., 2017). The Transformer architecture proposed by Google Research became the cornerstone of all modern LLMs (Brown et al., 2020), and models like GPT-3 started to demonstrate that the general capabilities of a model are closely related to its size (Kaplan et al., 2020). From ChatGPT, techniques like Reinforcement Learning from Human Feedback (RLHF) were used to "align" the model's behavior with human intent and values, greatly promoting the popularization of AI (Ouyang et al., 2022).

The emergence of powerful open-source models like LLaMA (Touvron et al., 2023), and QWen (Bai et al., 2023) has lowered the technical barrier. The current focus of development has shifted from pursuing larger parameters to pursuing better performance, especially in deep applications within vertical domains (Bommasani et al., 2021). Combining general-purpose LLMs with professional knowledge to create expert-level AI has become a clear industry direction (Zhao et al., 2023).

General-purpose LLMs, while knowledgeable, lack depth and precision in the rigorous medical field and carry the risk of "hallucination" (Ji et al., 2023). Healthcare is a necessity-driven market where users demand extremely high information accuracy (Topol, 2019). QWen-3 is a leading open-source model with hundreds of billions of parameters, demonstrating exceptional performance in Chinese understanding, logical reasoning, and coding (Yang et al., 2024). So we built our Erkang Diagnosis model based on the 30B QWen-3 model through specialization on our 500+GB of high-quality, proprietary medical data. By performing continued pre-training (Gururangan et al., 2020) and RAG (Lewis et al., 2020) on QWen-3, we can "inject" this knowledge into the model, forming a unique and difficult-to-replicate professional advantage. Simultaneously, we establish strict security and compliance filters on this foundation to ensure the model's outputs are always prudent and responsible, mitigating medical risks — something difficult to achieve quickly with a self-developed model (Zhang et al., 2023).

Our expected users do not need an AI that "can chat about anything," but rather an expert that is "trustworthy on health issues" (Davenport & Kalakota, 2019). The Erkang Diagnosis LLM interacts with the use 3-5 rounds of structured inquiry, and then provides personalized, evidence-based health assessments and recommendations. This not only solves user pain points like "difficulty accessing doctors and cumbersome consultations" but also builds brand trust through professional services, creating long-term value for the company.

## 2. The Erkang-Diagnosis-1.1 Model

In this section, we describe our model. In general, we adopt a hybrid architecture of "data dual-wheel drive", combining enhanced pre-training and retrieval-enhanced generation to ensure the balance between deep memory of knowledge and real-time accuracy.

The overall technical architecture diagram is shown in Fig. 1.

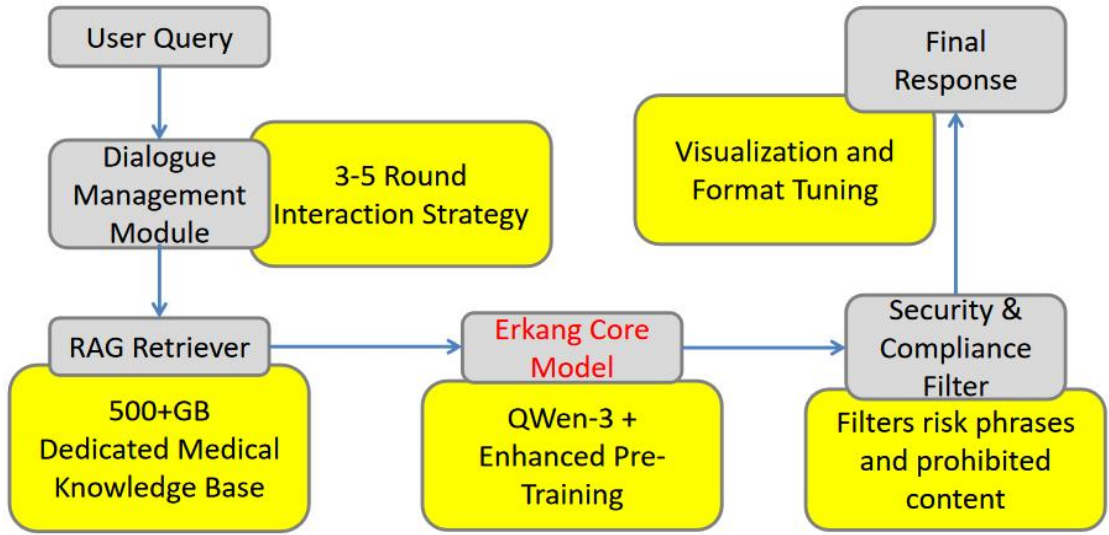

Fig. 1 The overall technical architecture diagram of the Erkang-Diagnosis-1.1 model

2.1 Base Model Selection: It's well-known that QWen-3 excels in general language understanding, logical reasoning, and instruction adherence, providing a solid foundation for domain-specific adaptations. It also has an excellent understanding of Chinese context and medical slang, in line with local user needs. So we use a 30B QWen-3 model as our base model.

2.2 Data engineering and knowledge fusion: We collected 500+GB of self-owned data, including:

Medical textbooks and guidelines: authoritative textbooks and clinical guidelines covering major departments such as internal medicine, surgery, gynecology and pediatrics.

Drug specification library: structured information on domestic and imported drugs.

Anonymized doctor-patient Q&A data: real consultation records that have been strictly anonymized and desensitized.

Medical literature and encyclopedia: high-quality medical paper abstracts, popular science articles and disease encyclopedia.

Medical knowledge graph from experts: medical knowledge graph on 200+ common diseases generated by renowned doctors.

We also did data preprocessing: Clean, deduplicate, and standardize raw data, and fragment knowledge to build high-quality training and retrieval corpora.

2.3 Model enhancement training strategy: To "internalize" 500GB of core medical knowledge into the model parameters, so that it has basic medical common sense and pathological reasoning ability, we used the continued pre-training technique, that is, the QWen-3 base model was fine-tuned with supervised learning on the preprocessed medical corpus, focusing on enhancing its capabilities in medical entity recognition, etiological chain reasoning, and diagnostic logic modeling. Subsequently, we use supervised fine-tuning with carefully constructed instruction samples simulating real 3-5 rounds of consultation scenarios, requiring the model to learn to actively ask questions about key symptoms (such as location, nature, duration, aggravating and relieving factors), perform differential diagnosis, and finally give prudent recommendations. More precisely, the 3-5 round interactive strategy engine is designed as a state machine to steer the conversation toward completing an effective consultation. Each round intelligently determines the next key symptoms to ask based on collected information (for example, if the user says "headache", the model will prioritize asking "What kind of pain is it?", "How long has it been?", and "Which part is the most painful?"). When sufficient information has been collected (typically 3-5 rounds) or when the situation is deemed urgent, the model will proactively conclude the inquiry and provide a structured summary and recommendations.

2.4 Retrieval Augmented Generation: Carefully selected medical knowledge, especially the medical knowledge graph from renowned doctors on 200+ common diseases, were built into a vector database using an advanced embedding model to convert text into vectors. Next, before the model generates the answer, the current user question is combined with the multi-round dialogue history, and the semantic retrieval is performed in the vector database to recall the most relevant medical knowledge fragments (such as disease description, medication guidelines, and examination recommendations), and the retrieved knowledge fragments are used as augmented context and input, together with the original dialogue history, into the tuned core model to ensure that its answers are based on the latest and most accurate factual basis.

2.5 Safety, Compliance, and Ethical Considerations: These are the lifelines of medical AI. We have established a multi-layered protection system: First, we have the content security filtering which integrates with multiple content filters. It strictly blocks requests involving self-harm, violence, illegal, and unethical content. Second, we have medical risk control: Disclaimer: At the end of each interaction, the model explicitly states that "I am not a doctor, and my advice is for reference only and cannot replace professional medical diagnosis." We also have the emergency identification and referral. That is, when the model identifies critical symptoms such as chest pain, acute abdominal pain, or impaired consciousness, it will immediately interrupt the conversation and strongly advise the user to "call 120 immediately or go to the nearest hospital emergency department" and we will show the user the contact information of the nearest emergency department. To be prudent, the model also tries to avoid absolute statements. Through instruction fine-tuning, the model is strictly restricted from using words such as

"confirmed" and "guaranteed cure", and replaced with prudent expressions such as "possible", "consider" and "suggest screening". Finally, we performed privacy protection that all user data are encrypted and anonymized during transmission and storage, and will never be used for any purpose other than model training and medical usage.

2.6 Visualization & Formatting Module: Before the final response is generated, we introduce a visualization and formatting module to ensure professional content is presented to the user in a clear, readable, and well-structured format. This module uses a rule engine and a keyword tagging system to format the model's raw text output. The output is automatically segmented into logical paragraphs such as "Symptom Analysis," "Possible Causes," "Recommended Actions," and "Urgent Notes.", and it uses bullet points to present lists of suggestions or symptoms. For keywords that we want users to notice, the module will mark them into blue color. These keywords include: high-risk and urgent keywords such as, `Seek immediate medical attention`, `Emergency`, `Chest Pain`, `Shortness of breath`, warning and cautionary keywords such as, `Recommend investigation`, `Be vigilant`, `Contraindication`, `Side effects`, etc. By the way, at the end of every response, we will add a disclaimer saying that: This AI analysis is for reference only and cannot replace professional medical diagnosis. Please seek immediate medical care in case of an emergency."

Here is an example of the Erkang Diagnosis model in Fig. 2.

Fig. 2 An interaction and output example of Erkang Diagnosis Model

## 3. Performance Evaluation and Verification

To validate our model, we have built a multi-dimensional evaluation system.

**Professional evaluation:** We invited a team of medical experts to build a test set covering 200+ common diseases. Considering the accuracy of diagnostic direction recommendations, completeness of key symptom inquiries, and compliance of medication recommendations, the experts found that on the internal test set, the model gave a preliminary diagnosis direction that was more than 90% consistent with the expert consensus.

**User experience evaluation:** We did a small-scale public beta testing and feedback collection considering the conversation fluency, question accuracy, and helpfulness rating. The vast majority of users said the model was "professional in its questions" and "clear and easy to understand", and could effectively solve the problem in 3-5 rounds.

More formally, we actually proposed a six-dimensional evaluation model to validate the effectiveness of a medical AI, which are:

Stylization: Responses to patients should include analysis of the condition, diagnostic and treatment recommendations, referral to appropriate medical departments, management suggestions, and personalized advice, ultimately providing users with comprehensive consultation.

Patient-Friendliness: Interact with users in an empathetic and patient-centered manner, fostering a sense of understanding and support. Demonstrate compassion and care to alleviate user anxiety, thereby enhancing trust and satisfaction.

Professionalism: Exhibit a high degree of professionalism, characterized by comprehensive medical knowledge and sound clinical judgment. Provide treatment recommendations and prescriptions that are not only accurate and reliable but also aligned with medical standards.

Safety: Ensure responses are safe and harmless, free from geographical or racial discrimination. Appropriately address unhealthy and unsafe issues, offering accurate and reliable advice to avoid misleading users or increasing health risks.

Fluency: Responses should be fluent, coherent, and logical to ensure readability and comprehension.

Proactiveness: When faced with unclear symptom descriptions or insufficient information, proactively ask users to provide relevant details and pose targeted follow-up questions.

The scores of our model on these six dimensions are shown in Fig. 3.

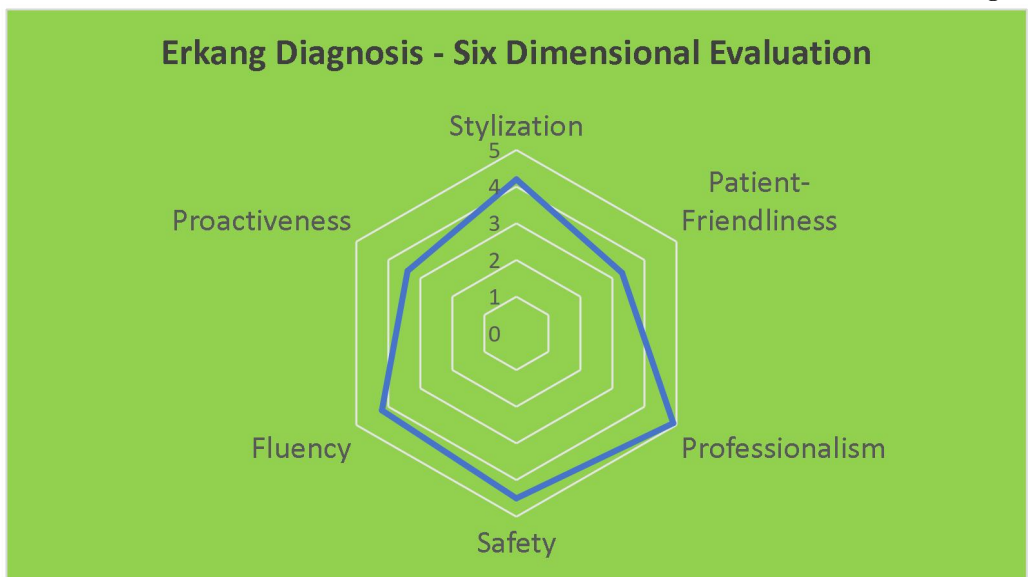

Fig. 3 The six dimensional evaluation of Erkang-Diagnosis Model

Moreover, the model was further compared with GPT-4 on its professionalism in the medical area, by means of taking the professional medical exams, as in Fig. 4.

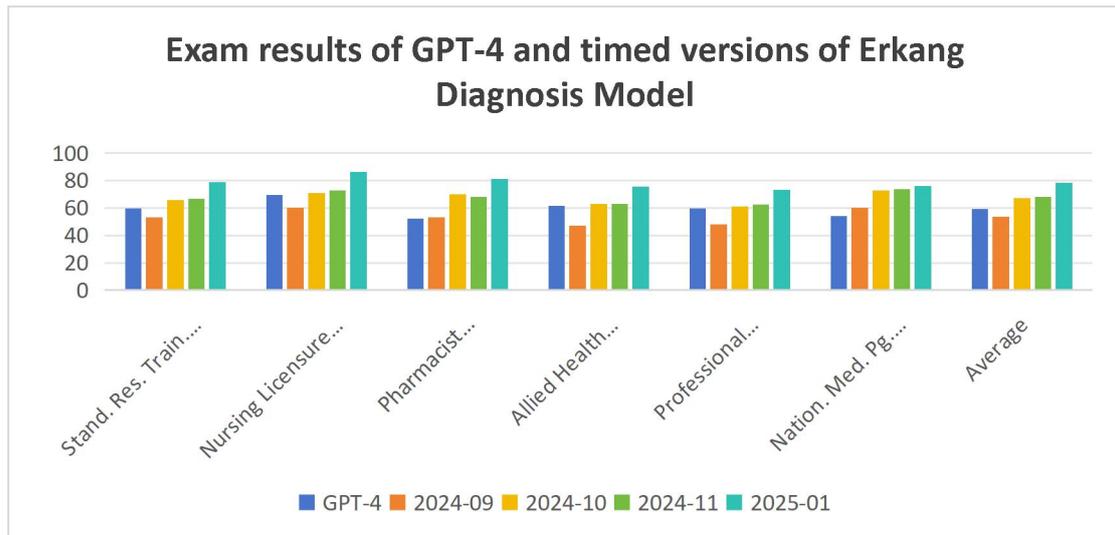

Fig. 4 The exam results of GPT-4 and versions of Erkang Diagnosis Model

## 4. Conclusion

In this work, we introduce Erkang-Diagnosis-1.1 which successfully integrates the general-purpose language model Qwen-3 with specialized medical knowledge, achieving a well-balanced combination of professionalism, security, and practicality through an innovative hybrid technical architecture. Using high-quality medical data and multistage pre-training and fine-tuning strategies, Erkang-Diagnosis-1.1 achieves excellent performance in providing medical advices.

In the future, we plan to integrate real-time medical literature and the latest clinical guidelines, and explore multimodal capabilities (such as interpreting skin and tongue coating images). We will explore the possibility of the model delving into specialized fields (such as mental health, nutrition, and rehabilitation) to develop more vertical consulting modules. Hospitals, physical examination centers and pharmaceutical enterprises can integrate the model capabilities into their service processes and create greater social and commercial value. We are confident that through continuous technological iteration and strict compliance control, the ErKang model will become an important force to promote the responsible application of AI in the medical and health field.